\documentclass[conference]{IEEEtran}
\IEEEoverridecommandlockouts
\usepackage{cite}
\usepackage{amsmath,amssymb,amsfonts}
\usepackage{algorithmic}
\usepackage{graphicx}
\usepackage{textcomp}
\usepackage{xcolor}

\usepackage{xcolor}
\usepackage{listings}
\usepackage{subfigure}
\usepackage{amsfonts}
\usepackage{url}
\usepackage{blindtext}

\def\BibTeX{{\rm B\kern-.05em{\sc i\kern-.025em b}\kern-.08em
    T\kern-.1667em\lower.7ex\hbox{E}\kern-.125emX}}
\begin{document}

\title{ ToriLLE: Learning Environment for Hand-to-Hand Combat }

\author{\IEEEauthorblockN{Anssi Kanervisto}
\IEEEauthorblockA{\textit{School of Computing} \\
\textit{University of Eastern Finland}\\
Joensuu, Finland \\
anssk@uef.fi}
\and
\IEEEauthorblockN{Ville Hautam\"aki}
\IEEEauthorblockA{\textit{School of Computing} \\
\textit{University of Eastern Finland}\\
Joensuu, Finland \\
villeh@uef.fi}
\thanks{This research was partially funded by the Academy of Finland (grant \#313970). We gratefully acknowledge the support of NVIDIA Corporation with the donation of the Titan Xp GPU used for this research. \

\copyright 2019 IEEE. Personal use of this material is permitted. Permission from IEEE must be obtained for all other uses, in any current or future media, including reprinting/republishing this material for advertising or promotional purposes, creating new collective works, for resale or redistribution to servers or lists, or reuse of any copyrighted component of this work in other works.}
}


\maketitle

\begin{abstract}
We present Toribash Learning Environment (ToriLLE), a learning environment for machine learning agents based on the video game Toribash. Toribash is a MuJoCo-like environment of two humanoid characters fighting each other hand-to-hand, controlled by changing actuation modes of the joints. Competitive nature of Toribash as well its focused domain provide a platform for evaluating self-play methods, and evaluating machine learning agents against human players. In this paper we describe the environment with ToriLLE's capabilities and limitations, and experimentally show its applicability as a learning environment with baseline and human experiments. The source code of the environment and conducted experiments can be found at \url{https://github.com/Miffyli/ToriLLE}.
\end{abstract}


\section{Introduction}
\label{intro}

Video games provide a rich and complex environments for training machine learning agents, without limiting them to real-time progression like with robotics. Popularity of such learning environments can be seen from the number of different video games used for such purpose, such as Atari games \cite{gym}, Doom \cite{vizdoom}, Quake \cite{dmlab} and Starcraft \cite{torchcraft,sc2}. These environments provide challenges for agents to complete in single-agent scenarios, which allows comparing performance of different learning methods, for example. At the same time there has been research towards super-human agents in various games, with success in mechanically simpler games like Chess and Go. However, when it comes to competitive video games, the complexity of the game itself is already a challenge in on itself.

For example, ViZDoom does support playing against other players, but learning to do so is difficult due to many problems environment presents \cite{vizdoom_competitions}: The agents must learn to navigate around map, explore options like shooting at enemies \textit{and} re-experience positive feedback many times for learning to happen. These challenges must be addressed before agents can start learning tactics against other opponents, who also try to out-smart their opponents. Especially training super-human agents in video games requires computer-cluster level of computing resources and manual tuning of the reward, actions and observations \cite{ctf,five,ssbm}. 

\begin{figure}[t]
\vskip 0.1in
\begin{center}
\centerline{\includegraphics[width=.75\columnwidth]{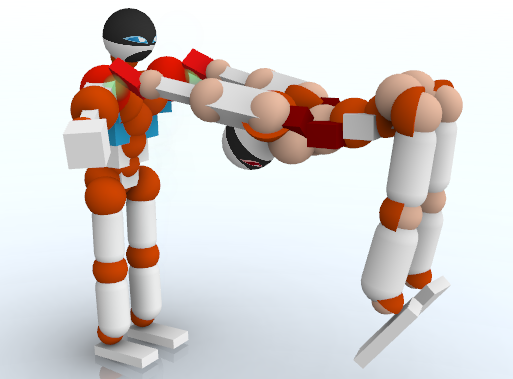}}
\caption{An in-game image of Toribash with two characters fighting each other. Two players control their respective characters by changing the state of the joints in the character's body. These states define how the joint behaves for the next simulated timesteps. Winner of the game is the one who received least amount of damage or the one who did not touch the ground with other than feet or hands.}
\label{im:toribash}
\end{center}
\vskip -0.2in
\end{figure}

For a more focused environment for competitive gameplay, we present a learning environment based on the video game Toribash \cite{toribash}, named Toribash Learning Environment, or ToriLLE.
Toribash is a game of two humanoid-characters competing against each other in martial-arts style and controlled by multiple individual joints in their bodies. Given its competitive gameplay mechanics, active player base of ranked human players and simple action/state space, Toribash ideal for studying how to learn robust policies against different opponents.

Toribash, and video games in general, is not designed for training machine learning opponents. It is designed for human players to be ran at comfortable frame rate for couple of hours at a time. To be a useful learning environment, the game should be fast and lightweight enough to support running multiple experiments and not crash during long experiments. To validate Toribash's applicability as a learning environment, we run baseline experiments with three reinforcement learning methods in the environment. To top this off we include self-play experiments and benchmarks against human players from the Toribash community. 

Although we present Toribash as a novel environment for machine learning, this is not the first time Toribash has been used in this context. A related publication used genetic algorithms to train an agent to attack an immobile opponent, and authors then analyzed how learned behaviour changed over time \cite{toribashga}. Outside academia, users from the Toribash forums have experimented with similar approaches in same task by using neuroevolution of augmented topologies \cite{toribashforum1,toribashforum2}. In fact, our work is inspired by and partly based on a code using genetic algorithms to successfully damage immobile opponent \cite{originalcode}. Our baseline experiments continue this trend by using recent reinforcement learning algorithms and models that depend on the current state of the game (i.e. see where the enemy is).

Our contribution can be summarized as such:
\begin{itemize}
    \item We present a new type of learning environment, especially suitable for studying reinforcement learning methods and competitive agents (e.g. self-play, game theory).
    \item We show the applicability of Toribash as a learning environment by running baseline experiments and studying game's resource use.
    \item We show that reinforcement learning agents and self-play can be used to reach performance that of a beginner human player in Toribash, without a large amount of training or modifications to existing learning algorithms.
\end{itemize}

\begin{figure*}[ht!]
\vskip 0.2in
\begin{center}
\centerline{\includegraphics[width=2.0\columnwidth]{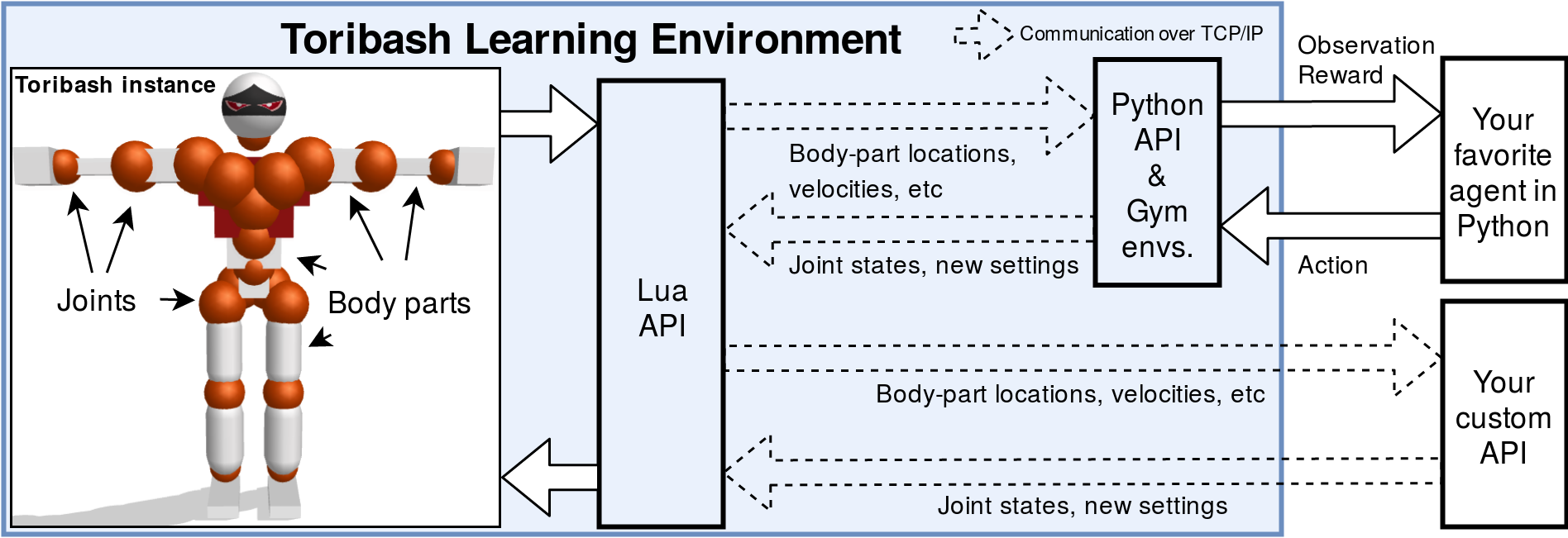}}
\caption{Overview of the Toribash Learning Environment (ToriLLE). ToriLLE uses Toribash's Lua scripts to communicate with outside controllers over TCP/IP. For each turn, game sends an observation vector to the controller. Controller replies to this by sending a list of joint states to be executed, and Toribash progresses by one turn.}
\label{im:torille}
\end{center}
\vskip -0.2in
\end{figure*}

\section{ToriLLE: Toribash Learning Environment}
\label{torille}

Here we briefly describe the video game Toribash, the ToriLLE environment and its performance as a learning environment.

\subsection{Video game ``Toribash"}
\label{sub:toribash}

    Toribash \cite{toribash} is a free game\footnote{We also include a copy of the game in our Python package.}, which consists of two humanoid characters playing against each other in martial-arts styled scenarios. See Figure~\ref{im:toribash} for an example. As of writing, there is an average of one hundred players online at any given time\footnote{Source: \url{https://steamcharts.com/app/248570} .}. Players have official ranking based on results of their games, similar to ELO-rating, with occasional tournaments where groups/clans or individuals compete against each other.
    
    Game progresses in turns. During each turn, both players set each of their character's joints to one of the four states: \textit{hold}, \textit{relax}, \textit{extend/raise} or \textit{contract/lower}. We will refer this as an \textit{action}. After both players have selected their actions, game progresses by a fixed number of frames, simulating the body movements according to the selected joint states. After this, next turn begins. Game ends when a set number of frames have passed or one of the players is disqualified, if such disqualifying rules are set. Joints are visible in Figures~\ref{im:toribash} and \ref{im:torille} as orange spheres. Detailed list of the joints and their possible states are listed in Table 1 of \cite{toribashga}.
    
    The game tracks amount of damage players' characters have received (we will call this \textit{injury}). Different parts of body receive different amounts of damage, e.g. punching the head will do more damage than punching the elbow. Game shows the injury of the opponent as the score of the player. Player with higher score, and thus least amount of injury, at the end of the game wins the game. Disqualification can be enabled as well, causing player who touches the ground with anything else than feet or hands to lose. Most of the human games end in such disqualification when disqualification is enabled\footnote{Private communicate with an experienced Toribash player.}. Players can also dismember characters: Hits strong enough will detach parts of the player from their bodies. The player is still able to control joints of severed limbs, and player is disqualified if such severed limb hits the ground.

\subsection{Why Toribash?}

    Toribash has the following positive highlights among other learning environments:
    
    \begin{itemize}
    \item[$+$] \textbf{Focused problem}: Toribash does not present difficulties like the need for exploration, multi-task learning or long-term temporal modelling, which is case with e.g. Dota 2 \cite{five} and Starcraft \cite{sc2}. We study this further in Section \ref{experiments}.
    \item[$+$] \textbf{Competition}: Being a competitive game, Toribash lends itself to game theory research. Specifically, Toribash calls for game theory techniques that also work in its high-dimensional and continuous state space. 
    \item[$+$] \textbf{Human players and replays}: Toribash has an active and ranked player base. This allows easy evaluation against top human players. The game allows replaying stored games, which can also be used for imitation learning using replays obtained from human players (e.g. on Toribash community forums\footnote{\url{http://forum.toribash.com/forumdisplay.php?f=10}.}).
    \item[$+$] \textbf{Turn-based}: Toribash runs in turns, much like Go and Chess, and players gain little benefit from fast reactions. This avoids problem of inhumane reaction-times being huge benefit over human players, which has been expressed as one the unfair advantages of AI in benchmarks like in Dota 2 \cite{five} and Super Smash Bros \cite{ssbm}.
    \end{itemize}
    
    But, in all fairness, Toribash has some limitations when it comes to using it as a learning environment: 
    
    \begin{itemize}
      \item[$-$] \textbf{Compatibility}: Toribash is developed originally for Windows using OpenGL. As such, running the game on Linux requires additional work (e.g. \texttt{Wine} library), and OpenGL requires a valid display to initialize correctly.
      \item[$-$] \textbf{Closed environment}: Toribash is not open-source and thus cannot be modified at a lower level.\footnote{However the main developer ``hampa" was kind enough to implement some required features for ToriLLE.}
    \end{itemize}
    
    Considering both the pros and cons, Toribash does not offer the flexibility and performance to be used as an environment for evaluating multiple algorithms against each other or in-depth analysis of learning methods. Environments like Atari Learning Environment, MuJoCo or Roboschool offer faster execution at a lower computational cost, as well as the possibility to create various environments and robots. The benefit of Toribash to research lies in its competitive nature in high-dimensional state space: Computer agents have to control a complex body (reinforcement learning, control), but also reason and plan ahead of the opponent (game theory).
    
    Other video-game environments like Starcraft \cite{sc2} offer similar challenges but require special learning methods and/or long training runs before learning to play against human players \cite{vinyals2019alphastar}. Meanwhile, current off-the-shelf reinforcement learning algorithms can be successfully trained on Toribash without additional modifications (Sections \ref{experiments} and \ref{selfplay-experiments}). We argue this simplicity makes Toribash ideal for studying how to combine reinforcement learning and game theory together.


\subsection{Playing Toribash via software}

    \begin{figure*}[ht]
    \begin{minipage}{.45\textwidth}
    \begin{lstlisting}[basicstyle=\small, language=Python]{Name}
    import torille
    from torille import utils
    toribash = torille.ToribashControl()
    toribash.init()
    while True:
        s,t = toribash.get_state()
        if t: break
        a = utils.create_random_actions()
        toribash.make_actions(a)
    toribash.close()
    \end{lstlisting}
    \end{minipage}
    \hspace{20pt}
    \begin{minipage}{.45\textwidth}
    \begin{lstlisting}[basicstyle=\small, language=Python]{Name}
    import gym
    # Registers environments
    from torille import envs
    e = gym.make("Toribash-DestroyUke-v0")
    while True:
        a = e.action_space.sample()
        s,t,r,i = e.step(a)
        if t: break
    e.close()
    \end{lstlisting}
    \end{minipage}
    \vspace{2pt}
    \caption{Example snippets of Python code running random agent on Toribash via ToriLLE, with default interface (left) and Gym environment (right). Settings of Toribash can be defined on creation and episode reset (not shown in the code).}
    \label{im:codes}
    \end{figure*}
    
    To train agents with Toribash we wish to play Toribash via programming languages directly. To this end, we use Toribash's Lua scripts to send the current state of the game and receive commands synchronously over TCP/IP, waiting for the actions from the controller (e.g. Python) before proceeding game forward. Controller can also control rules of the game like gravity and initial distance between players. 
    
    ToriLLE can be used with any language which supports TCP/IP networking. We provide a simplified library in Python which hides all the communication under an API similar to ViZDoom and OpenAI Gym \cite{gym}  environments. Figure~\ref{im:codes} provides examples of using direct API as well as basic example of the Gym API. Provided Gym environments implement required functions specified by OpenAI Gym documentation, which allows using ToriLLE environments as a drop-in replacement of other Gym environments.
    
    
    \textbf{For observation information}, ToriLLE provides 3D coordinates of the body parts' locations, their 3D velocities, rotation matrix of the groin (close to hip), states of joints, injuries of both players and winner of the game (if any). Body parts are visible as white shapes in Figures \ref{im:toribash} and \ref{im:torille}. This information does not describe the state of the game completely: We do not know the rotation of any dismembered limb, for example. The library can be modified to provide rotations of all joints and other objects in the future.
    
    The Python library includes normalization code for the observations, centering and rotating the coordinates of body parts relative to both character's groin. There are total of $21$ body parts on each characters' body, so e.g. positions of both characters would be a vector of length $21 \cdot 2 \cdot 3 = 126$. Toribash can also provide information on custom objects (e.g. size and location of obstacles), which can be supported by ToriLLE.
    
    
    \textbf{For actions}, ToriLLE takes in next state of each joint in characters' bodies. These states specify how the joint will actuate until next turn. There are total of $20$ joints for each character, each of which can be in four possible states $\{1,2,3,4\}$. Character also has hands which can also be in one of two states \textit{release} or \textit{grip}, which can be used to take hold onto opponent. 
    
    Upon end of the game the controller can change the rules of the game, change the game mod and/or enable recording of a replay of the next round. Controller can also read a stored replay to play it through and obtain per-frame information. Combined with e.g. human replays, this could be used for imitation learning or for learning the basic dynamics of the game.

\subsection{Performance}
\label{sub:performance}

    \begin{figure}[t]
        \centering
        \hspace{-0.1cm}
        \subfigure[Workstation (4-core 4.0GHz, Nvidia GTX 1080)]
        {
            \includegraphics[scale=0.42]{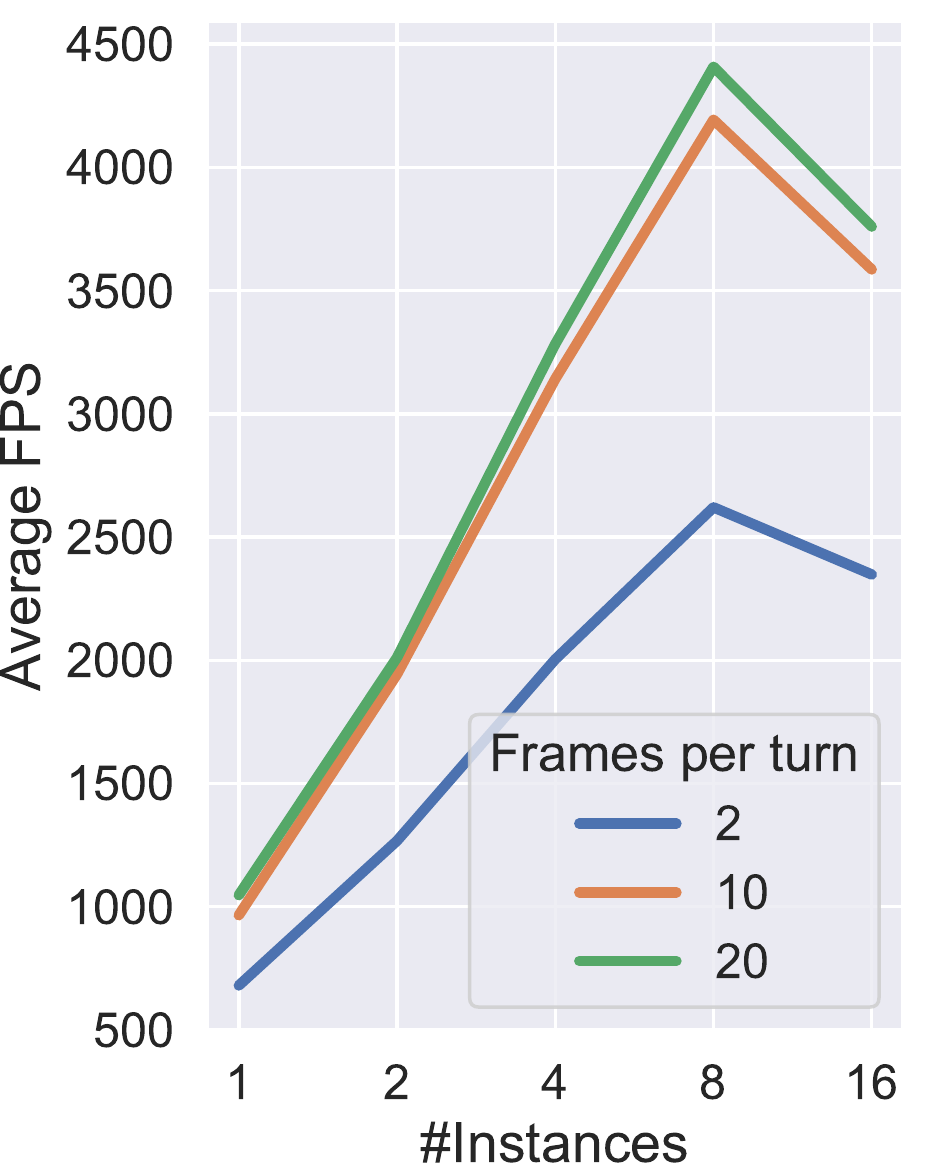}
            \label{im:workstation_perf}
        }
        ~
        \subfigure[Server (16-core 2.1GHz, Nvidia RTX 2080Ti)]
        {
            \includegraphics[scale=0.42]{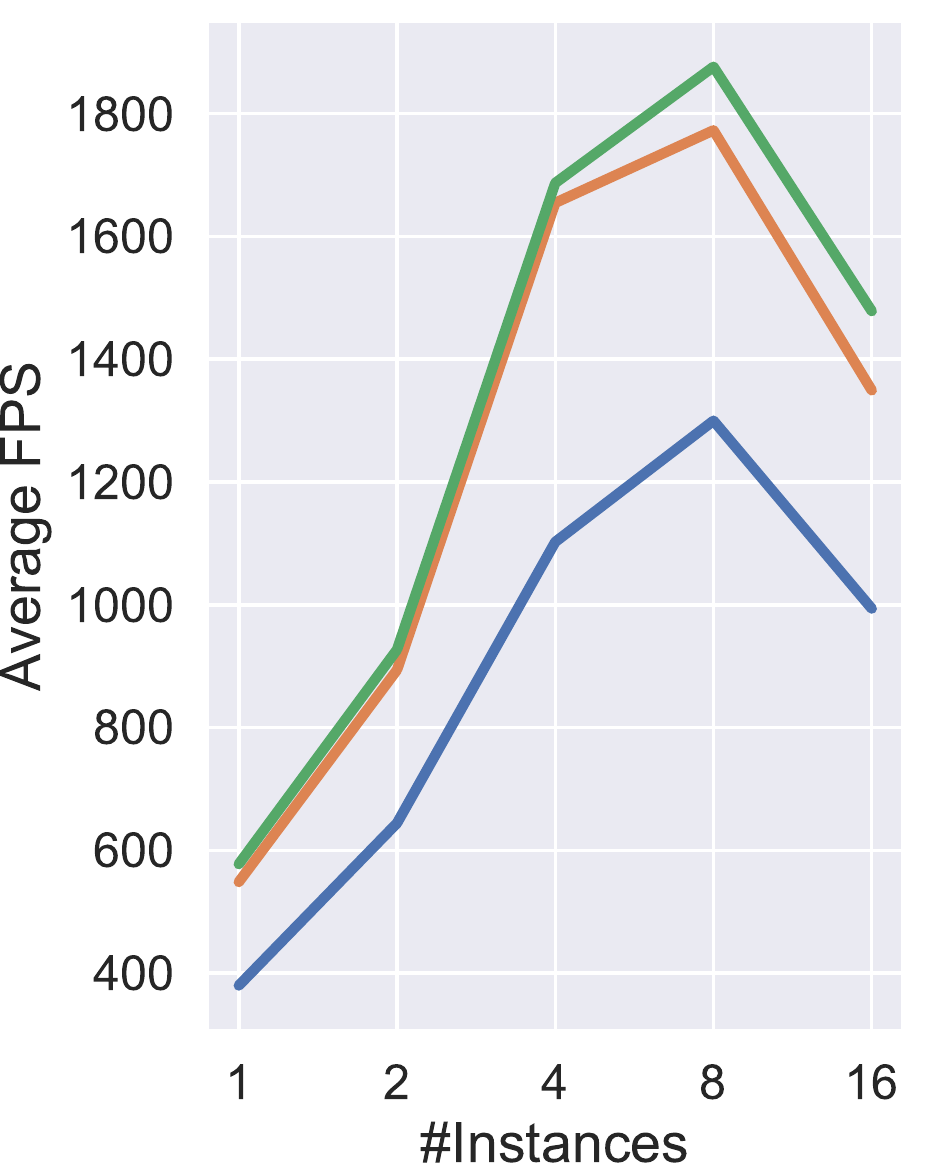}
            \label{im:server_perf}
        }
        \caption
        {
        Performance benchmark results of ToriLLE with episode length of $1000$ frames and engagement distance of $1500$ (i.e. no collisions between players). One agent step is after every ``Frames per turn" frames. Toribash can be ran at FPS comparable to Starcraft 2 environment \cite{sc2} and scales up to four parallel instances on same machine before benefits from more instances diminish. With higher frames-per-turn the game runs faster but also increases time between agent steps. Toribash runs notably faster on high clock-rate machines than on multi-core machines.
        }
        \label{im:benchmarks}
    \end{figure}

    Toribash is designed to be human-playable game at comfortable frame-rate of 60 frames per second (FPS), which is the designed pace of the game (frame-rate dictates simulation speed). However, being an older game designed to be ran on older hardware, we can run the game significantly faster on modern hardware. For a better picture, we ran performance benchmarks on two different machines.
    
    We ran benchmarks for a minute by executing random actions for both players and calculating the average FPS over the run. We varied the number of frames between turns to see the overhead caused by communication between Toribash and Python program. We also repeated experiments with multiple instances of ToriLLE in parallel asynchronously to see how well Toribash scales up for multiple cores. 
    
    Benchmarking results presented in Figure~\ref{im:benchmarks} indicate that Toribash scales up to four instances almost linearly, but beyond that the benefits begin to diminish. Curiously, even with $16$ physical CPU cores, moving from $8$ to $16$ parallel instances provides lower total FPS. We believe this has to do with varying clock-speeds of modern processors, where clock-speed of an individual core is reduced when processor is under a heavy load. The Toribash environments run notably faster on a quad-core, higher clock-speed CPU. With longer turns the overall FPS is higher but this also reduces the rate at which agent receives observations from the environment.
    
    The complexity of simulation also has influences the game speed: In our experiments in Section \ref{experiments} we noticed the speed of the environments changing as agent learns to punch the other player, requiring more computation to process the collisions. We do not include this effect in our benchmarks as it cannot be controlled with simple settings.
    
    While running and creating the benchmarks, we noted following three things related to the performance of the environment:
    
    \textbf{Toribash and Linux.} Up-to-date Toribash binary is only available for Windows on Steam, but modern versions of \texttt{Wine} are able to run Toribash on Linux machines as well. We used \texttt{Wine 3.0.2} to run the experiments included in this paper. 
    
    \textbf{Resources.} One instance of Toribash uses at most one CPU core, around 700MB of system memory and 20MB of GPU memory on Ubuntu 16.04. No memory leaking or slow-downs were observed during experiments done in this work. This allowed us to comfortably run $10$ instances of Toribash on a $16$-core server machine for a week.
    
    \textbf{Headless rendering.} Toribash uses OpenGL to render the game, which requires a valid display where screen buffer can be created. Creating valid screen buffers on headless servers and/or over SSH connections requires special setups. One way is to use virtual screen buffers like \texttt{Xvfb}. As of writing, Nvidia drivers do not work with \texttt{Xvfb}, and CPU-based rendering must be used. We observed notably smaller frame-rate from this setup, and especially the total FPS gain with multiple instances vanishes.

\section{Toribash as a machine learning environment}
    \label{experiments}
    
    \begin{figure}[t!]
        \centering
        \includegraphics[scale=0.5]{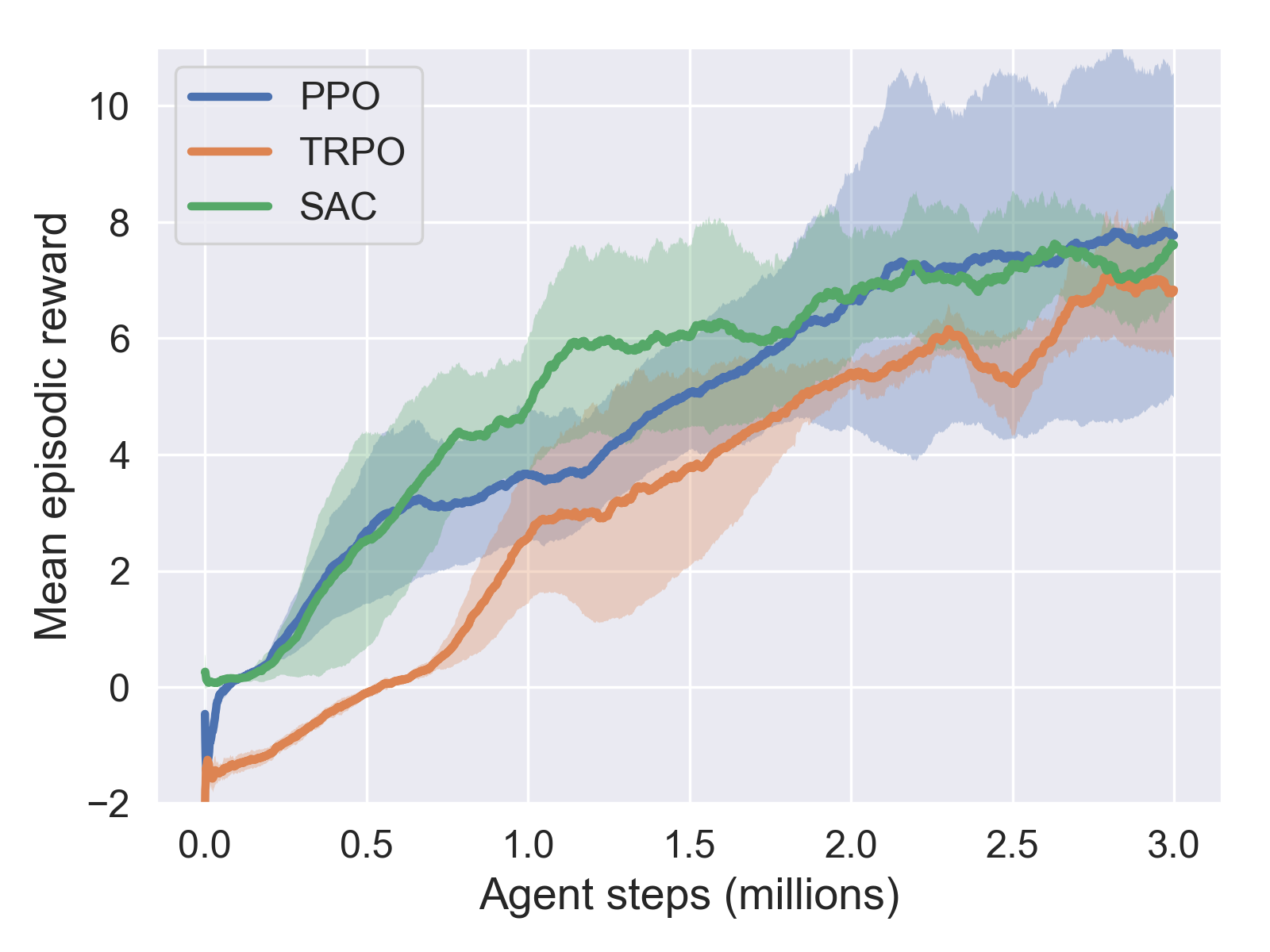}
        \caption
        {
            Learning curves of training three different learning algorithms with goal of damaging immobile target in Toribash. Each line is an average over five repetitions, and for each individual run one point on the curve is average performance over $200$ last games. Shaded area is the standard deviation over these five runs. Hyperparameters are selected per method for highest score. While these methods have been reported to have clear differences in performance in other experiments \cite{ppo, sac}, in Toribash their performance is approximately same. 
        }
        \label{im:results}
    \end{figure}
    
    Now for a pressing question: \textbf{Is this environment suitable for training machine learning agents?} Since the game is not originally designed to be used for experiments like this, there are no guarantees it would be fast or stable enough for long experiments spanning multiple days or even weeks. The game could be leaking memory and crash every now and then, or it could end up being too slow/heavy to run to obtain samples from thousands of games. 
    
    The environment can also prove to be too difficult to machine learning algorithms: Seemingly simple games like reaching goal in a small maze or Super Mario prove to be too difficult for current reinforcement learning algorithms by default, and they require additional modifications to learn successfully in the mentioned environments \cite{pathak2017curiosity}. If we want to focus on the competitive aspect of the game (i.e. game theory side), the environment mechanics and rules should be simple enough so that the learning algorithms can learn to play the game.
    
    To find answers to these questions we trained off-the-shelf reinforcement learning algorithms in this environment. By successfully running the experiments, we show that Toribash indeed is stable enough to be used for machine learning purposes. By training agents that improve over time in the task, we also show that the environment is simple enough to be learned by current algorithms without additional fine-tuning or modifications. 
    
    The code used to run the experiments is available at \url{https://github.com/Miffyli/ToriLLE/tree/master/experiments}. 

    \subsection{Environment and task}
        The environment consisted of the agent's character and \textit{Uke} character, which stays immobile. The task in the environment was to damage the Uke and avoid receiving damage yourself. The starting distance between players was randomly selected from the interval $[100, 200]$ to encourage more general solution rather than memorizing certain sequence of actions to play. The games lasted for $1000$ frames and each turn advanced by $10$ frames. This lower-than-normal number of frames per turn allows finer control over character than in default mods, where humans play with $20 - 40$ frames per turn.
        
        \textbf{Reward: } Given the change in player's injury between two successive turns $\Delta p$ and Uke's injury $\Delta u$, the reward signal for agent was $(\Delta u - \Delta p) / 5000$. That is: Damage the opponent without causing damage yourself. Note that this reward is always zero or positive, since amount of injury never decreases. The normalizing term was used included to avoid large magnitudes. Using logarithmic scale with appropriate base would also work as an normalizer for the reward, but accumulated reward in log-scale would be higher for many smaller hits than for few stronger hits (e.g. $4 \cdot \log_{10}(10)  > \log_{10}(1000)$).
        
        \textbf{Observations: }  Agent received normalized locations of both players w.r.t to player's groin, with rotation of player's groin included. We replaced the height coordinate of player's groin with the absolute height, so that the agent knew its relative position to the ground. The final observation space was thus $ x \in \mathbb{R} ^ {126}$, where values were clipped to interval $[-30, 30]$ to avoid too large numbers for the neural network.
        
        \textbf{Actions: } The action-space consisted of selecting one of four states for each joint, i.e. a multi-discrete variable. The agent chose one of four states for each of $22$ joints during each turn.
    
    \subsection{Learning algorithms}
        As per with reinforcement learning research, we wish to compare different learning algorithms by analyzing their performance in the environment over training regimen. Each training run lasted for three million agent steps (turns), and we repeated each experiment five times to include the variance between runs. 
        
        We selected three distinct actor-critic methods for our comparisons: \textit{Proximal Policy Optimization} (PPO) \cite{ppo}, \textit{Trust-Region Policy Optimization} (TRPO) \cite{schulman2015trust} and \textit{Soft-Actor Critic} (SAC) \cite{sac}. To avoid implementation bugs, we used pre-existing implementations of PPO and TRPO from stable-baselines package \cite{stable-baselines} and SAC implementation from rlkit package\footnote{\url{https://github.com/vitchyr/rlkit}.}.
        
        We selected these three based on their state-of-art results in MuJoCo locomotion tasks (similar to Toribash's character), for their applicability to multi-discrete action spaces (multiple joints with multiple discrete possibilities) and for existing implementations of the methods. Other possible algorithms include Branching DQN \cite{branchingdqn}, which modifies DQN to work with this type of action space, and evolution-strategy methods \cite{salimans2017evolution}.
        
        For all methods and experiments, we used a small network of two hidden layers with $64$ units each and tanh-activations. In PPO and TRPO experiments same network was used for estimating both value function and policy. In SAC experiments there were three separate networks: One for value estimation, one for state-action estimation (for computing advantage) and third network for policy estimation. All networks in all experiments were trained with an Adam optimizer \cite{adam}. 
        
    \subsection{Hyper-parameters}
        The three included RL learning algorithms are sensitive to settings set at the beginning of the training (``\textit{hyper-parameters}"), and Toribash presents a novel task to be completed, so we cannot be sure which set of hyper-parameters works best for Toribash tasks. For a fair comparison between these methods, we aim to find best hyper-parameters for each algorithm.
        
        To find suitable hyper-parameters, we trained learning algorithms on default \texttt{Toribash-DestroyUke-v1} Gym environment with different hyper-parameters. We then selected the best-performing hyper-parameters for the final experiments. In general, we found that amount of exploration (weight of entropy loss) plays a significant role across all three learning algorithms, and with too low or high weight the algorithms were unable to improve over time. We ran experiments over different magnitudes of entropy loss weight from in interval $[10^{-1}, 10^{-5}]$.  
        
        PPO is usually trained with multiple concurrent environments to gather more diverse dataset for updates, as well as with large batch size before updates. However, we found that PPO learned faster and scored higher with only one environment and batch size of $128$ experiences, compared to higher batch sizes or more environment. For weight of the entropy loss (encourages exploration) we select $10^{-4}$.
        
        TRPO is also trained with a single concurrent environment and $1024$ samples per update. Entropy loss weight $10^{-3}$ had highest score.
        
        While SAC also uses weight of the entropy loss to encourage exploration, the implementation also includes automatic entropy scaling \cite{haarnoja2018learning}. However, we found this automatic tuning to very quickly set the entropy weight very low, and agent was not able to improve after that point. Instead we select static entropy loss weight of $10^{-5}$. Reward-scale was mentioned as one of the sensitive parameters, and we found $1.0$ to perform best. Each policy update is done with $128$ samples from a single environment.
    
    \subsection{Experiment results}
        
        Starting with the question related to validity as a learning environment: All experiments including hyper-parameter search took a total of one week of computing on a 16-core server and quad-core workstation, and there were no issues observed (no random crashes or memory leaking). This shows that Toribash is indeed stable and fast enough to be used as a learning environment. Note that we only used one instances of the environment per experiment and ran multiple experiments in parallel. According to benchmarks in Figure~\ref{im:benchmarks}, one can also speed up individual experiment by running multiple environments per experiment. 
        
        As for the learning performance: Learning curves averaged over five repetitions are shown in Figure~\ref{im:results}. All of the three agents are indeed able to improve over time as training progresses, signaling that the task is indeed learnable without modifications to the learning algorithms. What is curious is the similar performance of all three learning algorithms: Research related to these methods shows that PPO is clearly outperformed by SAC in MuJoCo and robotics tasks \cite{haarnoja2018learning}, where controlled humanoids have similar structure. 
        
        The major difference between different learning algorithms is the variance of final performance between different runs: PPO's final performance varied in range $4.3$ to $12.7$, TRPO in range $5.8$ to $9.1$ and SAC in range $6.3$ to $8.7$. Upon subjective inspection of the learned models, TRPO and SAC learned very similar techniques across different runs despite the randomness of the environment (engagement distance) and different initial parameters. PPO learned different tactics across the different runs, sometimes ending up with just a strong kick which launched the opponent outside player's reach (lowest performance) but in one run agent learned to constantly pummel the opponent for more damage (highest performance).
        
        In light of these results, the requirement for reaching the optimal solution in this environment is not the best available learning method but a better exploration strategy: The main Achilles' heel of all tested learning algorithms was the fact that they initially learned some opening moves, and later on only built upon these without exploring different opening moves. This problem could be approached with novelty search / diverse policies \cite{eysenbach2018diversity}, where agents are encouraged to try novel approaches to the task.

\section{Self-play experiments and playing against humans}
\label{selfplay-experiments}
    Given the successful experiments with RL agents in the previous section, a natural follow-up question arises: \textbf{How well these agents could play against humans in Toribash?} Given the success of RL agents beating human players in various games \cite{alphago, five, ctf}, it indeed seems possible to do in Toribash as well. With this in mind, we continued by training RL agents with self-play to beat human players.
    
    \subsection{Setup}
        \textbf{Task and environment:} For game rules we selected a commonly played multiplayer mod \texttt{aikidobigdojo.tbm}. The mod includes a large area called \textit{dojo}. Players lose if they touch the ground with other parts than hands or feet inside dojo or with any part outside the dojo. Mod has time-limit of $500$ frames, and number of frames per turn increases from $10$ to $50$ over the game.
        
        \textbf{Observations:} Agents received relative locations of players' body-parts, rotated and translated according to their groin. With this information alone agent cannot know if it is inside the dojo, which is why we included player's distance from the center in the observations. We also included number of frames left in the game and number of frames for next turn. This observation does not include current joint states of either player, making the environment only partially observable. We compensated for this by using recurrent models. 
        
        \textbf{Agent:} We selected PPO algorithm based on the results from the previous experiment and due to its previous success in self-play experiments \cite{selfplay, five}. The neural network consisted of one $64$ unit layer followed by a $64$ unit LSTM layer \cite{gers1999learning}. With a LSTM layer, the agent is able to construct a hidden state that includes information from previous observations, e.g. model the velocity of the body parts and enemy behaviour. We set entropy coefficient to $0.01$ and number of samples per update to $512$. We continued using high-quality implementation of PPO from stable-baselines \cite{stable-baselines} in self-play experiments.
       
        \textbf{Self-play training:} PPO agent was trained by mostly playing against itself. We launched four separate instance of ToriLLE with above settings, in which the learning agent (one being updated) fought against different opponents. The opponent was either a randomly playing agent ($20\%$), one of the previous learned models ($20\%$) or the current learning agent ($60\%$). The opponent was randomly changed after every game with $1\%$ probability. We included random agents and previous versions to diversify the opponent pool, which has been found to be beneficial \cite{selfplay}. The agent being trained only played as the Player 1, while the opponent was always the Player 2. The game should be perfectly mirrored between the two players, and as such it should not matter as which player you play. However, following results show otherwise.
    
    \subsection{Training results}
        We ran two shorter experiments of 5M game-steps (run 1 and run 2) and one longer training of 20M steps (run 3, equal of a week of wall-clock time on a modern quad-core machine). The first two experiments did not include random agent as one of the opponents. The parameters of agent were saved at fixed intervals. To find a candidate to play against human players, we played each saved agent against all other saved agents for $100$ episodes. This provided a better picture which of the learned parameters was most promising to play against human players, as we compare one agent version against opponents they did not experience during training.
        
        \begin{figure}[t]
        \centering
        \includegraphics[scale=0.7]{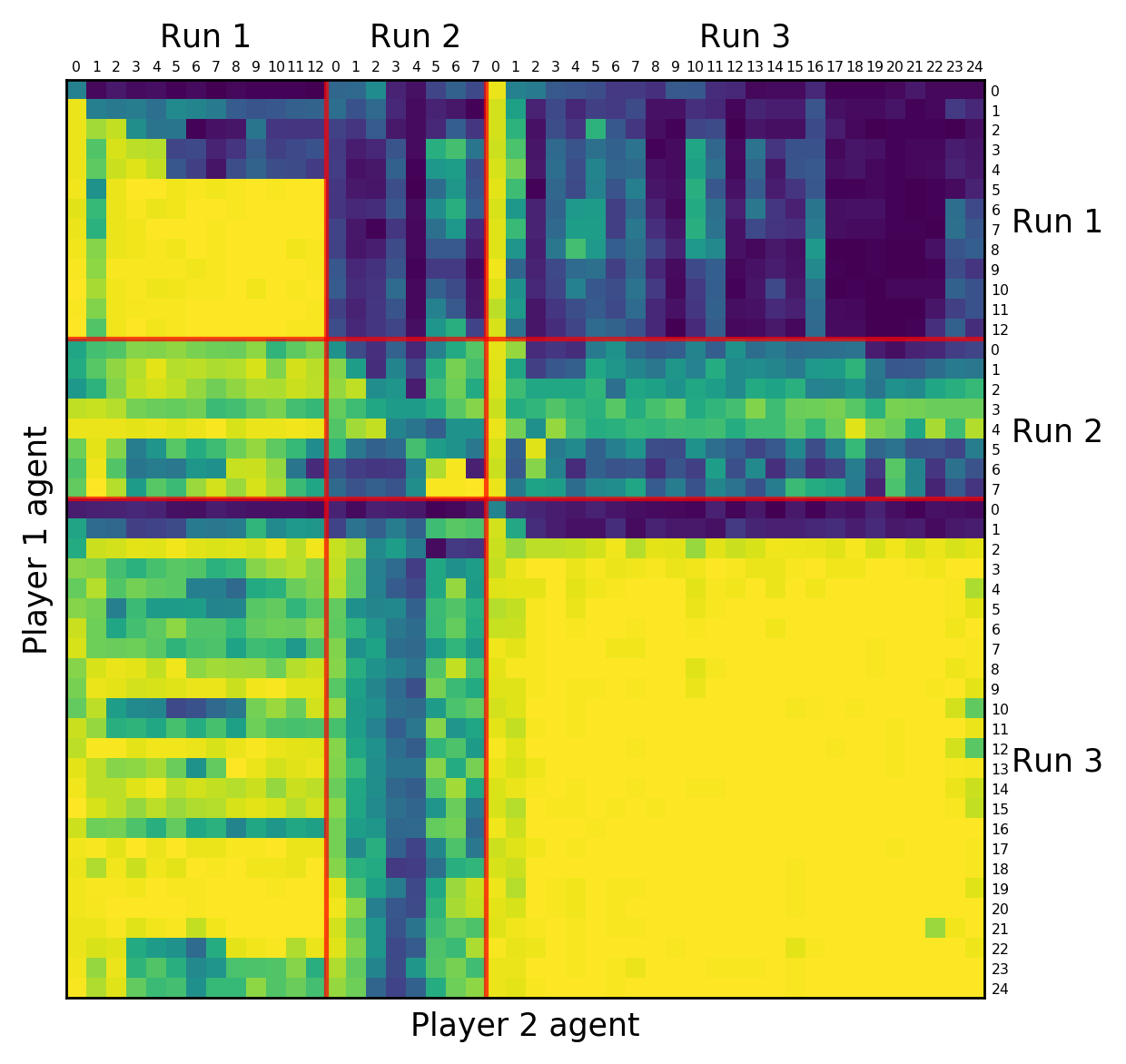}
        \caption
        {
            Performance of $46$ self-play agents from three different training runs playing against each other. Small numbers indicate agent's revision in that training run. Each pixel is agent on row as Player 1 versus agent on column as Player 2, and bright color represents higher score for Player 1. Ideally this matrix should look antisymmetric, but we see Player 1 often dominates regardless of the agent revision (run 1 and 3, bright boxes), even when both players are the same agent (bright diagonals). Between training runs the results are as expected (antisymmetric), with run 1 having weakest agents overall (dim upper right and upper center boxes).
        }
        \label{im:selfplay-results}
        \end{figure}
        
        The results of evaluating all saved agents against each other are presented in Figure~\ref{im:selfplay-results}. Note that each pair of agents is evaluated against each other twice but with flipped roles (Player 1 and Player 2). Ideally this matrix should look antisymmetric: If agent beats another as Player 1, it should do the same as Player 2. However, in the figure we see large areas where agent X wins over agent Y, but then agent Y wins agent X when player roles are flipped (bright boxes).
        
        Upon closer inspection of games between agents like this, we found that both players are receiving same observations and taking same actions initially. However, in the next turn Player 2 received a slightly different observation, with some variables being different from Player 1's by less than $1\%$. While a minimal difference, it was enough for Player 2 to pick different actions than Player 1, despite in reality the game state is mirrored between players. This difference in observations stems from rounding errors in observation handling (rotating by matrix multiplication), and possibly from the mechanics of Toribash (imprecision in the simulation). 
        
        This deviation from learned strategy as Player 2 led to Player 1 gaining upper hand constantly\footnote{Quite literally: Player 1 ripped off the arm of Player 2 and threw it into the ground, resulting in a victory for Player 1.}. This led to Player 1 being a constant winner in the game and learning algorithm enforced this behaviour, but this behaviour was not ideal for Player 2. This created a feedback loop of agent learning this one specific strategy which only works if the opponent does very same actions (which it did, thanks to self-play training). Use of older models of the agent and random opponents did not help with this issue.
        
        These results highlight the need of more diverse opponent pools, like having multiple different learning agents playing against each other \cite{ctf}. Especially when using self-play to train agents in Toribash environment, the agent should be exposed to playing as both players rather than just playing as Player 1 or Player 2, even if the observations are normalized to be the same for both players. 
        
    \subsection{Performance against human players}
        
        We selected version 24 of run 3 and version 4 of run 2 for their performance against all other agents, and then pit them against two human players from the Toribash community, one with several hundreds and another with more than two thousand hours of gameplay time. We did not gather personal information of the players for anonymity. Players played five rounds in total, three games against one of the models and then two rounds against the another, in random order. We did not disclose against which agent humans were playing against.
        
        The more experienced player won both models in all games. The other player won $1/3$ games against run 3 model, and $1/2$ of games against run 2 model. Both agents were said to have very defensive behaviour with focus on avoiding touching the floor, rather than actively trying to win the opponent. Both human players commented that both agents could be considered decent human players, and clearly above a novice player. The more experienced player placed them on same performance level as a human player with $20$ to $40$ hours of gameplay time.
        
        The behaviour was said to be deterministic as well, with little reaction to the current situation. Indeed, the actions taken by agent changed rarely as the game progressed or even against different players. The agents were also susceptible against out-of-training-set opponent tactics: Human players were able to throw the agent off the stage with no resistance from the agent, as the agent never saw this behaviour during its training.
        
        By successfully running these experiments we show that Toribash indeed can be used for self-play experiments along with human evaluations. While the results leave a lot to be desired in terms of performance against humans, it presents an exciting milestone among other video games used to benchmark autonomous agents against humans. 

\section{Conclusion}

    We presented ToriLLE: An agent learning environment based on video game Toribash, a humanoid martial-arts game.  Conducted experiments and benchmarks show Toribash's applicability as a learning environment. The prime contribution of Toribash to machine learning community is the focus on task of outplaying the opponent, such as human, in the game. Unlike other competitive learning environments like Starcraft 2, Toribash has simpler mechanics, which we show by successfully training off-the-shelf learning algorithms on the environment without per-environment modifications. We continued on these experiments with self-play learning in hopes of defeating human players, and the agent was able to learn beginner-level skills but was not able to win experienced human players. This task of defeating human players in Toribash thus remains an open problem and as an exciting field for future research. 
    
    As a related note, the next instalment of the Toribash, named ``Toribash Next", is under development and is implemented in Unity. Given Unity's support for machine learning agents \cite{unityml}, Toribash Next could lend itself more conveniently for machine learning purposes. 

\section*{Acknowledgements}
    We would like to thank Toribash developer ``hampa" for modifications to Toribash that made this project possible, and Aistis Ramonas for assistance with Toribash community and setting up the human evaluations. We thank Janne Karttunen and reviewers of previous versions of this article for constructive comments that helped us to improve this paper. 

\bibliographystyle{ieeetr}
\bibliography{main}

\vspace{12pt}

\end{document}